\pdfoutput=1
\documentclass[11pt]{article}
\usepackage{acl}
\usepackage{times}
\usepackage{latexsym}
\usepackage{booktabs}
\usepackage{array}
\usepackage[T1]{fontenc}
\usepackage[utf8]{inputenc}
\usepackage{microtype}
\usepackage{todonotes}
\usepackage{amsmath}
\usepackage{mathtools}

\title{Understanding Emotion Valence is a Joint Deep Learning Task}
\author{Gabriel Roccabruna, {\bf Seyed Mahed Mousavi}, {\textbf{Giuseppe Riccardi}} \\
         Signals and Interactive Systems Lab, University of Trento, Italy\\
         \texttt{gabriel.roccabruna@unitn.it,giuseppe.riccardi@unitn.it}}

\begin{document}
\maketitle
\begin{abstract}
The valence analysis of speakers' utterances or written posts helps to understand the activation and variations of the emotional state throughout the conversation. More recently, the concept of Emotion Carriers (EC) has been introduced to explain the emotion felt by the speaker and its manifestations. In this work, we investigate the natural inter-dependency of valence and ECs via a multi-task learning approach. We experiment with Pre-trained Language Models (PLM) for single-task, two-step, and joint settings for the valence and EC prediction tasks. We compare and evaluate the performance of generative (GPT-2) and discriminative (BERT) architectures in each setting. We observed that providing the ground truth label of one task improves the prediction performance of the models in the other task. We further observed that the discriminative model achieves the best trade-off of valence and EC prediction tasks in the joint prediction setting. As a result, we attain a single model that performs both tasks, thus, saving computation resources at training and inference times. 

\end{abstract}

\section{Introduction}
Speakers express their emotions in the language in different modalities (e.g. speech and/or text) and interaction contexts (e.g. dyadic interactions or social media posts). A type of document imbued with emotions conveyed through the recollection of personal events experienced by the speaker is the personal narrative. Personal Narratives (PN) have been recently studied to promote healthier mental health by modelling the patients' life events and monitoring emotional states \cite{mousavi2021unsupervised, danieli2021conversational, danieli2022assessing}.

\begin{figure}
    \centering
     \includegraphics[scale=0.62]{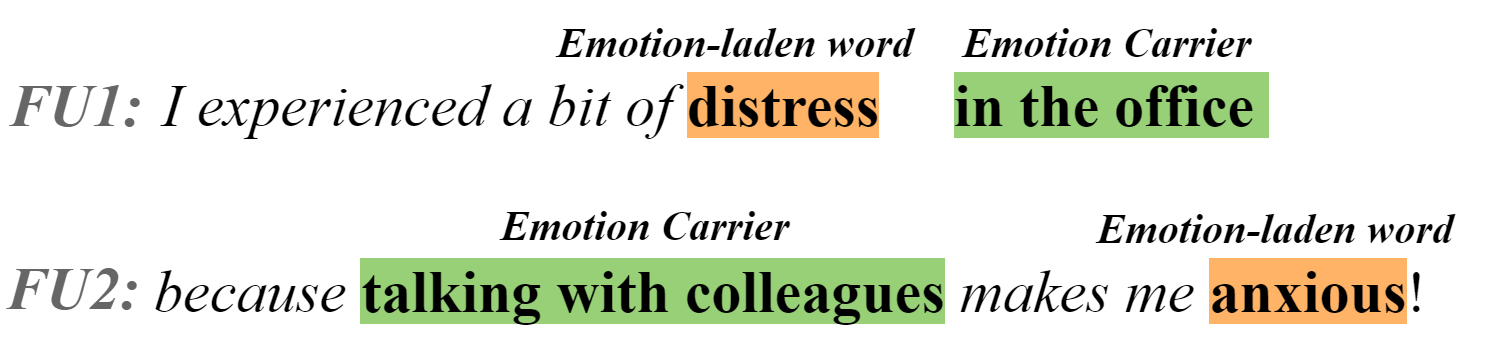}
    \caption{Example of two Functional Units (FU1, FU2) by \citet{mousavi-etal-2022-emotion}. In each unit, the emotion-laden words convey an explicit emotion while the emotion carriers are implicit manifestations of emotions even though they represent neutral emotion at the surface level.}
    \label{fig:wassaex}
\end{figure}

Monitoring the narrators' emotional states in PNs is achieved through valence analysis and the identification of related emotion carriers. Valence analysis addresses the identification of emotion levels ranging from pleasantness to unpleasantness generated by an event or a stimulus \cite{russell1980circumplex}. The valence can be manifested explicitly via emotion-laden words, such as \textit{Death} or \textit{Birthday} in the PN. Besides emotion-laden words, valence can also be manifested implicitly through Emotion Carriers (EC), i.e. persons, objects or actions, that may not represent any emotion at the surface level (such as ``the office'' or ``Wednesday''). Figure \ref{fig:wassaex} shows an example of a sentence consisting of two Functional Units (FU1, FU2; the minimal span of text expressing a dialogue act \cite{Bunt2012ISO2A}) by \citet{mousavi-etal-2022-emotion} with the emotion-laden words and the ECs in each unit. Recent studies show that ECs yield a detailed and understandable representation of the emotional state by highlighting the source of the valence such as ``colleagues'', ``a vacation'' or ``a stroll along the river'' \cite{tammewar2020annotation, mousavi-etal-2022-emotion}. 

\begin{figure*}[t]
    \centering
    \includegraphics[width=\textwidth]{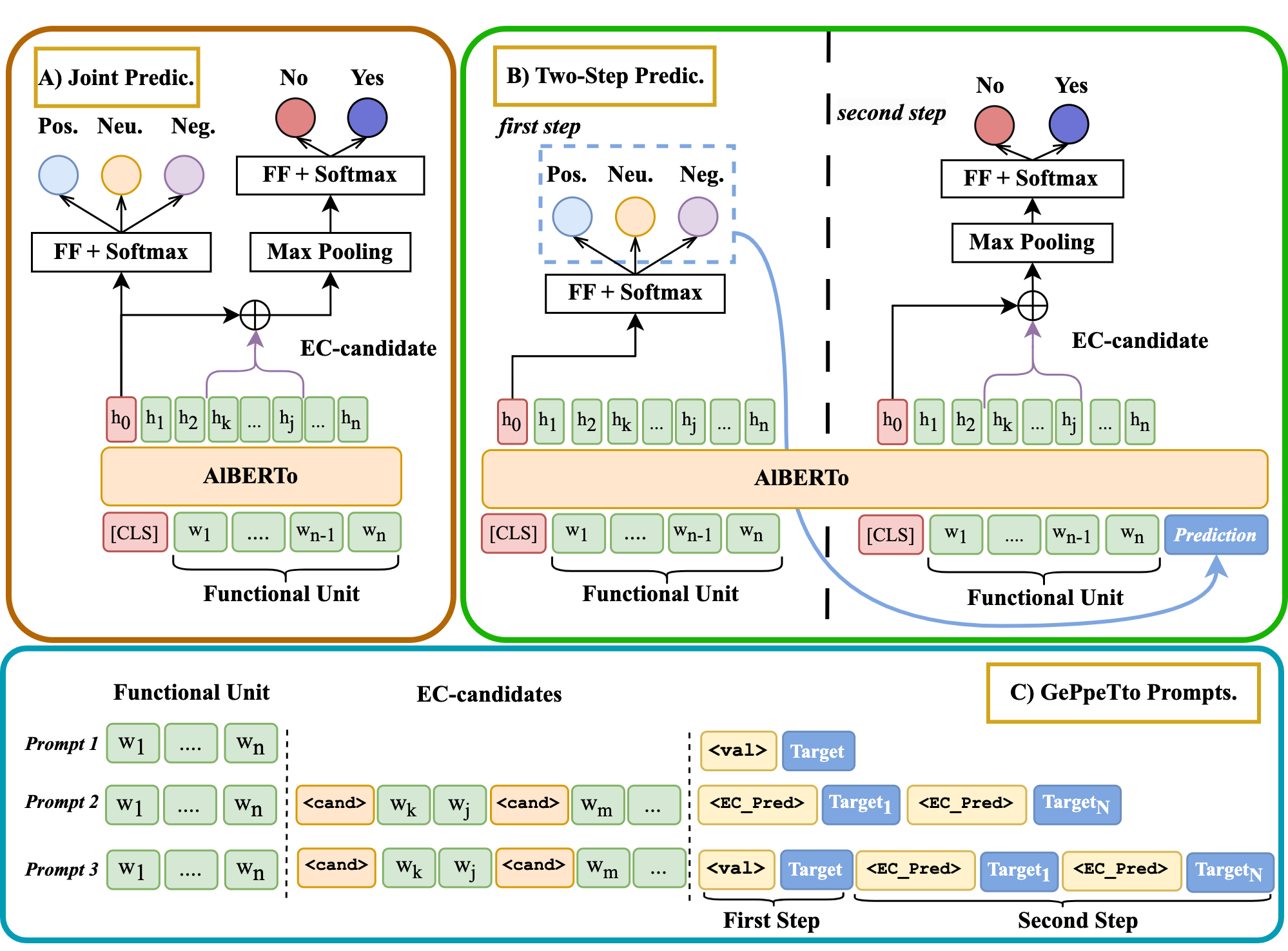}
    \caption{The joint and two-step settings applied to discriminative (AlBERTo) and generative (GePpeTto) PLMs in three sections: A) fine-tuning AlBERTo with the joint prediction; B) the two-step prediction applied to AlBERTo (the first task is valence prediction and the second task is EC prediction); C) the prompts designed to fine-tune GePpeTto for valence prediction (prompt 1), EC prediction (prompt 2), and two-step approach (prompt 3).}
    \label{fig:approaches}
\end{figure*}

The two elements of valence and EC are inter-dependant since valence represents the intensity of the experienced emotions while the ECs are the means through which emotions are expressed and conveyed throughout the PN. Consequently, when narrators recount an event that activated their emotional state, the intensity of such emotion is manifested as valence while the expression of the emotion is through the recollection of the event and/or the participants characterising the activation of the emotional state. 

In this work, we explore the natural inter-dependency of valence and the related ECs in PNs. This inter-dependency is characterised by the relations between the presence or absence of ECs and neutral or non-neutral valence. Namely, the presence of ECs in a FU implies a non-neutral valence, while neutral valence for a FU implies the absence of ECs. Moreover, the polarity of the valence might be related to the presence of domain-specific ECs. For instance, the ECs ``the office'' or ``boss'' might appear more frequently with a negative valence as opposed to ``vacation'' or ``children''.

We investigate the contribution of this inter-dependency in the prediction of the valence and the related ECs in a Multi-Task Learning (MTL) approach. MTL is to train a single model on multiple related tasks to achieve inductive transfer between the tasks, which is to leverage additional information sources while learning the current task. Inductive transfer enhances generalisation by introducing an additional source of inductive bias used by the learner to prefer a hypothesis over the other hypothesis \cite{caruana1998multitask}. We experiment with two MTL approaches i.e. joint and two-step prediction. While in the joint approach, the labels of the valence and EC prediction tasks are predicted simultaneously, in the two-step approach, the prediction of one label is used as the context for the prediction of the other task. 

We investigate whether this inter-dependency can be learned by Pre-trained Language Models (PLM). PLMs have prevailed over the other deep neural models in sentiment analysis \cite{Mao2021AJT, roccabruna-etal-2022-multi}, and they have been effectively used as a backbone of MTL architectures achieving state-of-the-art performance in intent classification and slot filling \cite{chen2019bert, Qin2020ACT}, dialogue state tracking \cite{hosseini2020simple, Su2021MultiTaskPF} and aspect-based sentiment analysis \cite{Mao2021AJT, jing2021seeking}. 

We experiment with discriminative AlBERTo (BERT) as well as generative GePpeTto (GPT-2) models. In particular, for the discriminative model, we combine the two architectural solutions for valence and EC prediction tasks proposed by \citet{mousavi-etal-2022-emotion}. While for the generative model, we design two prompts for valence and EC prediction and one prompt for the two-step prediction setting. In these experiments, we evaluate both models for each MTL setting, where the baseline is the performance of the model fine-tuned on every task separately. Moreover, we compute the upper bound in the two-step prediction setting by replacing the first prediction with the ground truth. This upper bound also represents the level of inter-dependency between the two tasks. In this work, we use a corpus of PNs in Italian with valence and ECs annotated at the functional unit level.


The contributions of this paper can be summarized as follows:

\begin{itemize}
    \item We study the inter-dependency of the valence and related Emotion Carriers in the corresponding prediction tasks;
    \item We fine-tune two PLMs and experiment with multi-task learning settings for valence and Emotion Carrier predictions;
    \item We evaluate and compare the performance of discriminative and generative models in the task of valence and Emotion Carrier prediction. 
    
\end{itemize}

\section{Related Works}

\textbf{Valence \& Sentiment Analysis} The values of valence have been studied both in a continuous space \cite{ong2019modeling, kossaifi2017}, and discrete space with a Likert scale \cite{tammewar2022annotation,mousavi-etal-2022-emotion}, ranging from negative (unpleasant) to positive (pleasant). Using the discrete approach, valence can be assessed with different levels of granularity as the narrative level and functional unit level. A Functional Unit (FU) is the minimal span of text expressing a dialogue act \cite{Bunt2012ISO2A, roccabruna2020multifunctional}. Narrative-level valence analysis provides a general yet flat perspective of the narrators' emotional state \cite{Schuller2018}, meanwhile, the sentence-level and FU-level analysis provide a detailed perspective as it highlights the variations and fluctuations of the valence throughout the narrative \cite{mousavi-etal-2022-emotion}.

A common practice in developing models for emotion analysis is to model valence analysis as sentiment analysis by mapping the valence values into three sentiment classes, i.e. \textit{positive}, \textit{negative}, and \textit{neutral} \cite{roccabruna-etal-2022-multi, mousavi-etal-2022-emotion}. However, valence differs from sentiment as the latter identifies the polarity of attitudes or beliefs, such as \textit{hating} or \textit{liking}, towards a person (e.g. a politician) or an object (e.g. product or a movie) \cite{scherer2000psychological}. Meanwhile, valence represents the level of emotions in such as \textit{anger} or \textit{happiness}. 

\textbf{Emotion Carrier and Valence:} Emotion carriers are closely related to emotional valence as they explain the valence. \citet{mousavi-etal-2022-emotion} studied the correlation between the sequence tokens and the predicted valence (sentiment). The authors observed that the model focuses more on emotion-laden words (explicit), whereas humans identify the emotion carriers to explain the valence.

\textbf{Multi-task Learning:} Multi-Task Learning (MTL) has been used for affective computing in aspect-based sentiment analysis \cite{Schmitt2018JointAA, Mao2021AJT, jing2021seeking}, and emotion classification and emotion cause extraction \cite{turcan2021multi}. MTL has been studied using discriminative models (BERT) for entity-relation extraction\cite{xue2019fine}, as well as generative models (GPT-2) for task-oriented dialogues \cite{hosseini2020simple}, dialogue state tracking \cite{zhao2021effective}, and task-oriented response generation \cite{su2022multi}.

\section{Approach}
\label{sec:task_formalization}
The inter-dependency of the valence and ECs results in the co-occurrence of both elements in the same utterance as the valence represents the level of emotion and the ECs are expressions through which the emotion is conveyed. In other words, the neutral valence of an utterance implies the absence of ECs, while the presence of ECs indicates a valence polarity for the same utterance. Furthermore, the ECs in an utterance can provide insights into the polarity of the utterance valence since certain ECs are more often associated with negative emotions (such as deadline) or positive ones (such as graduation).

We investigate the natural inter-dependency of valence and ECs via the MTL approach, which allows the models to leverage additional information learned from other tasks while learning the current task, improving the generalization. We experiment with joint training, where the two labels of valence and ECs are predicted simultaneously, and two-step prediction, where one of the labels is predicted and used by the model to condition and guide the prediction of the second label. 

\subsection{Dataset} We use a corpus of written PNs collected and annotated by \citet{mousavi-etal-2022-emotion}. The dataset consists of 481 narratives from 45 subjects, who were employees with stress, with valence and EC annotations at the Functional Unit (FU). Out of 4273 FUs in the narratives, 40\% are annotated by polarity (13\% positive and 27\% negative) and the related ECs, while the rest are annotated as neutral and do not contain any EC. Considering both neutral and non-neutral FUs, 18.5\% of the span candidates are annotated as ECs (over 10763 span candidates). While the number of ECs considering only non-neutral FUs is 44.7\% over 4452 span candidates. We use the official splits of the dataset, provided by the authors, consisting of train (80\%), validation (10\%) and test (10\%) set, stratified on the polarity distribution.

To measure the number of ECs specific for a valence polarity, we started by computing the intersection of ECs set annotated in FUs with positive and negative valence. We observed that only 4\% of the ECs (14.8\% from the positive and 6\% from the negative sets) are present in FUs with both polarities and can convey both positive and negative emotions. That is, the majority of ECs (the remaining 94\%) are indicators and carriers of only one valence polarity. Table \ref{tab:examples} presents a representative sample of the ECs extracted from the three sets. 

\begin{table}[]
    \centering
    \begin{tabular}{c|c|c}
        \textbf{Positive} & \textbf{Negative} & \textbf{Intersection}\\
        \hline
        \textit{``perfume''} & ``to tackle'' & ``work''\\
        \textit{``vacations''} & ``administration'' & ``home''\\
        \textit{``yoga'' }& ``dentist'' & ``lunch''\\
        \textit{``a stroll''} & ``be late'' & ``today''\\
        \textit{``freedom''} & ``charged with'' & ``feeling''\\         
    \end{tabular}
    \caption{Examples of Emotion Carriers (EC) in positive and negative functional units of the dataset used (English translations). ``Intersection'' consists of the ECs that are present in both positive and negative FUs.}
    \label{tab:examples}
\end{table}

\subsection{Multi-Task Learning}
We experiment with Multi-Task Learning (MTL) approach to exploit the dependency between the valence and EC prediction tasks. We compare the performance of the models for EC and valence predictions as task-specific models, as well as two-step and joint-prediction models, and evaluate the performance.

\textbf{Single-Task Prediction} In the single-task prediction, the models are trained and optimized for each task separately. This modality is a baseline to compare model performance in other MTL settings such as two-step and joint prediction.

\textbf{Two-step Prediction} Inspired by \citet{kulhanek-etal-2021-augpt} and \citet{hosseini2020simple}, we experiment with the two-step prediction setting. In this setting, the model predicts the discrete label for the first task as the first step, and as the second step, this prediction is concatenated to the input sequence following a prompt structure to predict the label for the second task via the same model. Afterwards, the loss values of the two tasks are summed or aggregated with a linear interpolation before back-propagating it. In this setting, we experiment with alternating the order for the two tasks (Valence $\rightarrow$ EC, vs. EC $\rightarrow$ Valence).  

The motivations behind this setting are that 1) the contribution of one task over the other task is explicit, enhancing the understanding of the inter-dependency between the two tasks; 2) this approach can potentially reduce the gap in the performance between two interdependent tasks by conditioning the prediction of the second task with the prediction of the first task. In this, the best-performing task is placed in the first step. To provide evidence of this, we experimented by replacing the first step with an oracle providing the ground truth. 

The two-step setting is similar to the pipeline setting \cite{zhang-weiss-2016-stack} and Stack-Propagation framework \cite{zhang-weiss-2016-stack, qin-etal-2019-stack}. The two-step prediction is similar to the pipeline setting because the discrete output of a task is explicitly used in the prediction of another task, but in the pipeline setting, two different models are utilised instead of one. Moreover, the two-step prediction is close to the Stack-Propagation framework as the back-propagation of the loss updates the weights of the model used to predict the first task as well as the second task. However, the label guiding the model's prediction of the second task is not differentiable as in the Stack-Propagation. 

\textbf{Joint Prediction} The joint prediction setting is commonly used in MTL \cite{cerisara-etal-2018-multi, jing2021seeking} where a single model predicts labels for the different tasks simultaneously. Thus, the prediction of one task does not explicitly contribute to the prediction of another task. The loss is computed and back-propagated as in the two-step prediction approach.

\section{Models}

We experiment with discriminative (BERT) and generative (GPT-2) models and investigate the performance of the two models for the joint and two-step prediction of valence and ECs. The joint and two-step prediction settings along with the two architectures are depicted in Figure \ref{fig:approaches}. All the hyperparameters and model settings are reported in Appendix \ref{sec:appendix} to achieve the reproducibility of the results. 

\subsection{Prediction tasks}

In the discriminative model, we formalize the valence and Emotion Carrier prediction tasks as text classification tasks by following the formalization of \citet{mousavi-etal-2022-emotion}. The valence prediction task is formally defined as calculating the probability for a given functional unit as $p(valY_i| FU_i)$, where label $valY_i \in \{positive, negative, neutral\}$ and $FU_i = \{w_1, w_2, .., w_n\}$ as a sequence \textit{i} tokens $w$. Meanwhile, the EC prediction task is to predict for each EC candidate span, which is an automatically extracted verb or noun chunk, in a FU if it is an EC or not. That is $p(ecY_j| EC\text{-}candidate_j, FU_i)$ where the $FU$ provides context information for the prediction, $EC\text{-}candidate_j = \{w_k,..,w_l\} \in FU$ and $ecY_j \in \{yes, no\}$. 

In the generative model, inspired by \cite{hosseini2020simple}, we model the two tasks as causal language modelling tasks, in which the model is tasked to learn the joint probability over a sequence. For valence prediction, the sequence used to train the model is formally defined as $x_i = [FU_i; valY_i]$, i.e. the concatenation of $FU_i$ and $valY_i$, where the functional unit $FU_i$ is the context for the model in the prediction of $valY_i \in \{positive, negative, neutral\}$.  While the training sequence for the EC prediction task is $x_i = [FU_i; EC\text{-}candidate_i; ecY_i]$ where $FU_i$ is the functional, $EC\text{-}candidate_i$ is the complete list of the EC-candidate spans of $FU_i$, and $ecY_i$ is the list of the EC decision labels, i.e. $\{yes, no\}$, corresponding to the list EC-candidate span. In this sequence, both $FU_i$ and $EC\text{-}candidate_i$ is the context for the model in the prediction of $ecY_i$.

\subsection{Discriminative} 
Discriminative models based on PLMs have been effectively used for text classification tasks \cite{lei2019bert}, however, such models may need additional architectural components, such as conditional random fields and/or additional feed-forward layers \cite{shang-etal-2021-span}, to tackle a specific task. 

\textbf{\textit{Architecture}} Our discriminative model is based on the same architectural components for valence and EC predictions proposed by \citet{mousavi-etal-2022-emotion}. This architecture is composed of a PLM and a set of feed-forward layers used to make the prediction. The PLM is based on AlBERTo which is BERT-based with 110M parameters pre-trained on a corpus in the Italian language collected from Twitter \cite{polignano2019alberto}. The PLM takes as input a FU  with special tokens {\fontfamily{qcr}\selectfont[CLS]} and {\fontfamily{qcr}\selectfont[SEP]}, added at the head and the tail of the FU, and returns a sequence of hidden states.  The valence is predicted from the hidden state of the {\fontfamily{qcr}\selectfont[CLS]} token by first applying a feed-forward layer with softmax to compute the probabilities over the classes of the valence (\textit{positive}, \textit{negative} and \textit{neutral}). For the EC prediction, an EC-candidate span is represented by a set of hidden states corresponding to the tokens of the span. Furthermore, the hidden state of the {\fontfamily{qcr}\selectfont[CLS]} token is concatenated to the hidden states of the EC span  to give context information contained in the FU. These hidden states are passed through a max-pooling layer, to get the vector representation of the EC candidate, and a feed-forward layer with softmax to yield the prediction on the two classes (\textit{yes} and \textit{no}).

\textbf{\textit{Joint}} Regarding the join prediction setting, as depicted in Figure \ref{fig:approaches} part \textit{A)}, the valence and ECs are predicted in one step. In this, the shared part of the model between the two tasks is the PLM AlBERTo only. 

\textbf{\textit{Two-step}} The model in the two-step prediction setting, Figure \ref{fig:approaches} part \textit{B)}, has the same shared parts of the joint model, but the prediction of valence and ECs are done in two steps. The prediction of the first task is computed on a FU, while the prediction for the second task is computed on the concatenation of the FU with the label predicted in the first task. The prompt that concatenates the prediction when the first task is valence prediction is: 
$$FU_i = \{w_1, w_2, .., w_n\}$$
$$FU_i \oplus \textit{valence:} \oplus Prediction_i$$ 
where $i$ is a functional unit of the dataset, $\oplus$ is the concatenation by white space, $\textit{valence:}$ is plain text, $Prediction_i=(0|1|2)$, and $\{0,1,2\}$ are the labels \textit{negative}, \textit{positive} and \textit{neutral} respectively.  When the first task is EC prediction the prompt is 
$$ EC_j = \{w_k,.., w_l\} \in FU_i $$
$$ FU_i \oplus \textit{EC:} \oplus \{EC_1,.., EC_N\}$$

where $\textit{EC:}$ is plain text, and $\{EC_1,.., EC_N\}$ is the list of EC spans of the detected ECs in the $FU_i$.  Furthermore, to reduce the training time and stabilise the performance, we experiment with the teacher forcing technique \cite{lamb2016professor} that substitutes the prediction of the first task with the ground truth with a certain probability (to be selected as a hyperparameter). 

\textbf{\textit{Loss function}} In both joint and two-step prediction settings, the loss function is the cross entropy and the loss values of the two tasks are combined with a linear interpolation:
$$ loss_{total} = \lambda * (loss_{valence})+ (1- \lambda) * loss_{EC}$$
where $\lambda$ is a hyperparameter with a range from 0 to 1.

\subsection{Generative}

We used GePpeTto \cite{de2020geppetto} an auto-regressive model based on GPT-2 architecture which is pre-trained for the Italian language with 117M parameters. For valence, EC and two-step predictions, we have designed three prompts by following the formalization of the two tasks. 

\textbf{\textit{Prompt design}} The prompt for valence prediction is composed of two segments, where the first segment is the $FU_i$ and the second segment is the valence label $valY_i$ to predict preceded by a special token. This prompt is depicted in Figure \ref{fig:approaches} part \textit{C)} \textit{prompt 1}, where $Target=(0|1|2)$ indicating \textit{negative}, \textit{positive} and \textit{neutral} respectively. The prompt used for EC prediction is organised into three segments: a) $FU_i$; b) $EC\text{-}candidate_i$ spans separated by a special token and; c) the list of labels corresponding to each EC candidate $ecY_i$ separated by another different special token. This prompt is shown in Figure \ref{fig:approaches} part \textit{C)} \textit{prompt 2} where $Target = (y|n)$. A difference with the discriminative model is that in EC prediction the predicted label of one EC candidate is used as context to predict the next EC candidates due to the fact that the model is auto-regressive.  

\textbf{\textit{Joint}} In the joint prediction setting, we fine-tune a single model on valence and EC predictions using the two corresponding distinct prompts appearing in the same training batch. Thus, the prediction of one task does not occur in the context of the other task. 

\textbf{\textit{Two-step}} For the two-step prediction, we designed a specific prompt by combining the prompts for valence and EC predictions, which is composed of the 4 segments: a) $FU_i$, b) $EC\text{-}candidate_i$ spans, c) valence $valY_i$  and d) ECs $ecY_i$ targets, Figure \ref{fig:approaches} part \textit{C)} \textit{prompt 3}. The first two segments are the $FU_i$ and $EC\text{-}candidate_i$. The other two segments are the target labels of the valence and EC predictions tasks. Thus, in the first step, the model predicts, based on $FU_i$ and $EC\text{-}candidate_i$, the labels of the first task that are used as context in the second step to predict the other task. Moreover, alternating the order of the last two segments results in two prediction settings Valence $\rightarrow$ EC and EC $\rightarrow$ Valence.

\textbf{\textit{Generation strategy}} At inference time, the generation of the target is guided by forcing the special tokens, i.e. they are not predicted by the model, into the generated sequence and limiting the possible output labels by considering the probabilities of the tokens in our searching space i.e. $\{0,1,2\}$ for valence prediction and $\{y,n\}$ for the EC prediction. Moreover, for the EC prediction task, we force the same number of special tokens of the EC candidates to get one output label for each EC candidate, relieving the model from the complexity of counting the EC candidates.  

\textbf{\textit{Loss function}} The generative model is trained as a language model, i.e. the model is tasked to predict the next most probable word given a sequence of words. In this, the loss function is the cross entropy with the objective of minimizing the perplexity on the training set. 

\section{Experiments}
\label{sec:RA}

\begin{table*}[]
\centering
\small
\renewcommand{\arraystretch}{1.2} 
    \begin{tabular}{p{2.5cm}c c ccc}
        \multicolumn{6}{c}{\textbf{Valence Prediction}} \\
        \toprule
         \textbf{Model} & \textbf{Single Task} &  \multicolumn{3}{c}{\textbf{Two-Step}} & \textbf{Joint}\\
         \cline{3-5}
         &   & Val. $\rightarrow$ EC & EC $\rightarrow$ Val. & w. ground truth &\\
          \hline
         AlBERTo & 76.0 & 76.0 & 75.7  & 81.2 & 76.0\\
        \hline
         GePpeTto & 77.1 & 74.7 & 65.1 & 86.5 & 75.6 \\
        \cline{2-6}
          \hspace{3mm}+ domain adapt. &  - & 77.0 &  70.6 & -  &-\\
\hline \hline
        \multicolumn{6}{c}{\textbf{EC Prediction}} \\
        \toprule
         AlBERTo & 63.7 & 63.4 & 64.8 & 74.9& 65.0 \\
        \hline
         GePpeTto & 57.8 & 58.3 & 58.2 & 65.4 & 59.5 \\
         \cline{2-6}
         \hspace{3mm} + domain adapt. &  - & 59.5 & 60.7 & - &   -\\
         \bottomrule
    \end{tabular}
    \caption{The macro-F1 scores (average over 10 runs) in percentage of the Valence Prediction and Emotion Carrier (EC) Prediction tasks. The scores are achieved using discriminative (AlBERTo) and generative (GePpeTto) PLMs with single-task, two-step and joint prediction settings. Single task and two-step with (w.) ground truth are respectively the baseline and the upper-bound for the joint and the two-step settings.}
\label{tab:data_statistics}
\end{table*}

Table \ref{tab:data_statistics} presents the macro F1-score achieved by AlBERTo and GePpeTto models with single-task, two-step and joint prediction settings. 

The results on valence prediction achieved via the single-task discriminative model are on-par with those reported by \citet{mousavi-etal-2022-emotion}, while the results achieved on EC prediction are incomparable with \citet{mousavi-etal-2022-emotion} since our training set consists of all FUs with the authors train the models only using the FU with a valence polarity.

Regarding the single task setting, we observe that AlBERTo outperforms GePeTto for EC prediction, while GePpeTto outperforms AlBERTo on the valence prediction task. 

Regarding the two-step prediction setting Val $ \rightarrow $ EC, we observe a slight worsening in the performance of EC prediction for the AlBERTo model and a drop in valence prediction score for the generative model compared to the single-task setting. Nevertheless, domain adaptation improves the performance to achieve close results on the valence prediction and boosts the performance on EC prediction. Domain adaptation is performed by initially fine-tuning only on the first task and further fine-tuning on both tasks with the two-step approach. 

Regarding the reverse order of predictions, i.e. EC $ \rightarrow $  Val., the models do not manage to outperform the single-task alternatives on valence prediction, with the degradation being more significant for the generative model. On the contrary, the models exhibit a better performance for EC prediction in the two-step setting compared to single-task models, where GePpeTto with domain adoption achieves its best performance in all settings. 

Additionally, we have computed the upper bound for the two-step prediction by substituting the prediction of the first step with the corresponding ground truth and fine-tuning the model only on the second step. The results show a solid contribution of one task in predicting the other task by outperforming the models in all the other prediction settings. Furthermore, GePpeTto and AlBERTo achieve the highest performance in valence prediction and EC prediction respectively.

The results of the discriminative model in the joint prediction setting are on par and better than the other two settings for valence and EC predictions, respectively. While the performance of the generative model is worse than the two-step prediction for both tasks, but slightly better than the single task in EC prediction. In particular, AlBERTo achieves the highest macro F1-score on the EC prediction task compared to the other settings and the generative model. 

\section{Discussion}
The inter-dependency between valence prediction and EC prediction tasks is quantified in terms of performance by the upper bound computed by substituting the first step prediction with the ground truth. We observe that the performance of both tasks is enhanced.  Furthermore, we observe that the proposed MTL approaches, i.e. two-step and joint prediction, are effective in exploiting such inter-dependency. In particular, the two-step prediction boosts the performance of the generative model, while joint prediction improves the performance of the discriminative model. Moreover, we observe that these improvements affect mainly the EC prediction task. This is because, compared to valence prediction, the EC prediction task is objectively more challenging for the models due to the unbalanced distribution and the sparsity of the ECs (some ECs are personal w.r.t the narrator). Thus, the predictions of the EC tasks are too noisy to be exploited by the models to improve the performance of valence prediction. Indeed, the worst performance for valence prediction is achieved by both models in the two-step setting in which the first task is the EC prediction (i.e. EC $ \rightarrow $  Val.).

Regarding the comparison between discriminative and generative models, the best trade-off between valence and EC prediction tasks is achieved by AlBERTo, although GePpeTto fine-tuned with the single-task setting achieves the best performance on valence prediction.


\section{Conclusions}
\label{sec:conclusion}
In this work, we studied the inter-dependency between valence and ECs in personal narratives. For valence and EC prediction tasks, we compared task-specific models with two MTL settings, namely joint and two-step prediction. We experimented with discriminative and generative PLMs. The results indicate that PLMs fine-tuned with MTL settings achieve improved performance by exploiting the inter-dependency between valence and EC prediction tasks. In particular, the two-step setting is more effective for the generative model, while the joint setting best fits the discriminative model. Furthermore, the generative model outperforms the discriminative model on the valence prediction task, while the discriminative model achieves better results on EC prediction and the best trade-off between the valence and EC prediction tasks. Consecutively, one discriminative model performs the two tasks, reducing the demand for computational resources at training and inference time and, therefore, lowering carbon emissions in the environment.  

\section{Future works}
\label{sec:f_w}
In the two-step prediction setting, we have only experimented with unidirectional inter-dependency of the two tasks i.e. EC $\rightarrow$ Val. and Val. $\rightarrow$ EC. A possible future work is to design and experiment with a neural network that bidirectionally exploits the two predictions implementing the configuration Val. $\xleftrightarrow{}$ EC. However, a larger dataset with more narratives per narrator is needed as a positive contribution of ECs to the Valence prediction task is observed on \textit{with ground truth} setting only due to the fact that ECs are sparse in the corpus and specific w.r.t the narrators. 


\section*{Limitations}
\label{sec:limitations}
The dataset used in this work is in Italian and the PLMs are pre-trained for the Italian language. The performance of the models and the results may be influenced by language-specific properties.

To reduce the ECs sparsity and, therefore, better modelling the inter-dependency between EC and Valence prediction tasks, particularly in the experiments EC $\rightarrow$ Val., a larger dataset with more narratives per narrator is needed.

\section*{Acknowledgement}
We acknowledge the support of the MUR PNRR project FAIR - Future AI Research (PE00000013) funded by the NextGenerationEU and the PNRR Project ``Interconnected Nord-Est Innovation Ecosystem (iNEST)'', ECS00000043.

\bibliography{anthology,custom}
\bibliographystyle{acl_natbib}


\onecolumn 
\appendix
\section*{Appendix}
\subsection*{A Hyperparameters} 
\label{sec:appendix}
\begin{table*}[t]
\centering
\small
\renewcommand{\arraystretch}{1.2} 
    \begin{tabular}{cc cc ccc}
        \toprule
         \textbf{Model} &  \textbf{Parameter} & \multicolumn{2}{c}{\textbf{Single Task } }& \multicolumn{2}{c}{\textbf{Two-Step}} & \textbf{Joint}\\
         \cline{3-4}
         \cline{5-6}
         &   & Valence Pred. & EC Pred. & Val. $\rightarrow$ EC & EC $\rightarrow$ Val. &\\
          \hline
         AlBERTo & Learning Rate & 5e-5 & 4e-5 & 4e-5  & 6e-5 & 1e-5\\
        \hline
         AlBERTo & $\lambda$ & - & -& 0.5 & 0.4 & 0.3\\
        \hline
         GePpeTto & Learning Rate & 9e-3 & 8e-3 & 9e-4 & 7e-4 &  8e-3\\
         \bottomrule
    \end{tabular}
    \caption{List of hyperparameters used to fine-tune the two models.}
\label{tab:hyperparams}
\end{table*}

The special tokens used in the prompts, \texttt{<val>}, \texttt{<cand>} and \texttt{<EC\_pred>}, are added to the vocabulary of the model. Moreover, we encoded valence textual labels with numbers because they are language-independent and perform better than additional special tokens.

We used AdamW \cite{loshchilovdecoupled} as an optimization algorithm to fine-tune the discriminative and generative models. To stabilise the performance while fine-tuning, we used a linear warm-up scheduler on the learning rate with the warm-up steps set at 10\% of the total training steps \cite{mosbach2021stability}. We used the library Optuna optimizer \cite{optuna} to search for the best hyperparameter for each setting (single prediction, joint prediction and two-step prediction) and models (discriminative and generative). The complete list of learning rates is presented in Table \ref{tab:hyperparams}. The learning rates used in the two-step with ground truth are the same as  Val. $\rightarrow$ EC when the first step is valence prediction and  EC $\rightarrow$ Val. when the first task is EC prediction. Moreover, we used a batch size of 32 for both models, 30 epochs for AlBERTo and 60 epochs for GePpeTto and, early stopping with patience set to 5 epochs. In the two-step experiments, we used a teacher forcing probability of 1.0 in Val. $\rightarrow$ EC and 0.1 in EC $\rightarrow$ Val. . We trained our models using one single 3090Ti GPU.

\end{document}